\documentclass[journal]{IEEEtran}
\usepackage{cite}
\usepackage{amsmath,amssymb,amsfonts}
\usepackage{algorithmic}
\usepackage{graphicx}
\usepackage{textcomp}
\usepackage{xcolor}
\usepackage{underscore}
\usepackage{listings}
\usepackage{xcolor}
\usepackage{tcolorbox}
\usepackage{framed} 
\usepackage{multirow}
\usepackage{enumerate}
\usepackage{hyperref}
\usepackage[ruled,linesnumbered]{algorithm2e}
\ifCLASSINFOpdf
\else
\fi

\begin{document}
\title{BenchENAS: A Benchmarking Platform for Evolutionary Neural Architecture Search}

\author{Xiangning Xie,
		Yuqiao Liu,~\IEEEmembership{Student Member,~IEEE},
        Yanan Sun,~\IEEEmembership{Member,~IEEE},\\
        Gary G. Yen,~\IEEEmembership{Fellow,~IEEE},     
        Bing Xue,~\IEEEmembership{Senior,~IEEE},
        Mengjie Zhang,~\IEEEmembership{Fellow,~IEEE}

\thanks{X. Xie, Y. Liu, and Y. Sun are with the College of Computer Science, Sichuan University, Chengdu 610065, China (e-mail:xiangningxie99@gmail.com;lyqguitar@gmail.com;ysun@scu.edu.cn).}

\thanks{G. Yen is with the School of Electrical and Computer Engineering, Oklahoma State University, Stillwater, OK, USA (e-mail:gyen@okstate.edu).}

\thanks{B. Xue and M. Zhang are with the School of Engineering and Computer Science, Victoria University of Wellington, Wellington, New Zealand (e-mails:bing.xue@ecs.vuw.ac.nz;mengjie.zhang@ecs.vuw.ac.nz).}

}

\maketitle

\begin{abstract}
Neural architecture search (NAS), which automatically designs the architectures of deep neural networks, has achieved breakthrough success over many applications in the past few years. Among different classes of NAS methods, evolutionary computation based NAS (ENAS) methods have recently gained much attention.
Unfortunately, the issues of fair comparisons and efficient evaluations have hindered the development of ENAS.
The current benchmark architecture datasets designed for fair comparisons only provide the datasets, not the ENAS algorithms or the platform to run the algorithms. 
The existing efficient evaluation methods are either not suitable for the population-based ENAS algorithm or are too complex to use.
This paper develops a platform named BenchENAS to address these issues.
BenchENAS aims to achieve fair comparisons by running different algorithms in the same environment and with the same settings.
To achieve efficient evaluation in a common lab environment, BenchENAS designs a parallel component and a cache component with high maintainability.
Furthermore, BenchENAS is easy to install and highly configurable and modular, which brings benefits in good usability and easy extensibility.
The paper conducts efficient comparison experiments on eight ENAS algorithms with high GPU utilization on this platform. The experiments validate that the fair comparison issue does exist, and BenchENAS can alleviate this issue. A website has been built to promote BenchENAS at \url{https://benchenas.com}, where interested researchers can obtain the source code and document of BenchENAS for free.
\end{abstract}

\IEEEpeerreviewmaketitle

\section{Introduction}
\IEEEPARstart{D}{eep} Neural Networks (DNNs), as the cornerstone of deep learning techniques~\cite{lecun2015deep}, have achieved remarkable success in many real-world applications, such as classifying objects from images~\cite{2016Deep},~\cite{huang2017densely}, reasoning natural languages from texts~\cite{devlin2018bert}, and recognizing speech from voice signals~\cite{zhang2017very}, to name a few.
The vast successes of DNNs are generally credited due to the design of novel DNN architectures. This can be evidenced from VGG~\cite{simonyan2014very}, ResNet~\cite{2016Deep}, DenseNet~\cite{huang2017densely}, and Transformer~\cite{vaswani2017attention}, which have significantly different neural architectures from each other. Commonly, such novel architectures of DNNs are often manually designed with rich expertise~\cite{simonyan2014very,2016Deep,huang2017densely}. However, with the increasing number of the state-of-the-art DNN building blocks, such as inception modules~\cite{szegedy2015going}, residual connections~\cite{2016Deep}, or dense connections~\cite{huang2017densely} integrated into a high-performance DNN architecture, it is increasingly difficult to handcraft DNN architectures with exceptional performance. Furthermore, with the remarkable performance of DNNs, more and more researchers who are not experts of DNNs are trying to explore the amazing functionality of DNNs for the task at hand. Unfortunately, due to their limit or even without knowledge in designing promising DNN architectures, it is hard to obtain DNNs with satisfactory performance. This urgent demand has promoted the feverish development of NAS techniques~\cite{zoph2016neural}, which aims to automatically generate robust and well-performing DNN architectures by formulating as optimization problems and then solved via well-designed optimization algorithms.
 
Based on the optimization algorithms adopted, existing NAS algorithms can be broadly classified into three different categories: Reinforcement Learning (RL)~\cite{kaelbling1996reinforcement} based NAS algorithms~\cite{zoph2016neural}, gradient-based NAS algorithms~\cite{liu2018darts}, and Evolutionary Computation (EC)~\cite{back1997handbook} based NAS algorithms (in short named ENAS)~\cite{real2017large}. 
Specifically, the RL-based NAS algorithms often consume heavy computational resources due largely to the inexact reward resulted from RL techniques~\cite{sun2019evolving}.
The gradient-based algorithms are more efficient than the RL-based algorithms. Unfortunately, the gradient-based algorithms require constructing a supernet in advance, which also demands specialized expertise. 
The ENAS algorithms solve the NAS problems by exploiting Evolutionary Computation (EC) techniques. Specifically, EC is a class of computational paradigms such as Genetic Algorithms (GA)~\cite{mitchell1998introduction}, Genetic Programming (GP)~\cite{banzhaf1998genetic}, and Particle Swarm Optimization (PSO)~\cite{kennedy1995particle} solving challenging optimization problems by simulating the evolution of biology or swarming social behavior~\cite{sun2018igd,deb2002fast,sun2018improved,jiang2017transfer,sun2017reference}. In EC, an initial population of candidate solutions is generated and iteratively updated by the evolutionary operators for generations. Each individual in each generation needs to be evaluated to obtain the fitness value. As a result, the population will gradually evolve to increase in fitness to arrive at the best solution.
Compared to the RL-based NAS algorithms, ENAS algorithms often require less computation budget. 
Compared to the gradient-based NAS algorithms, ENAS algorithms are often fully automatic, and they can produce an appropriate NAS without any human intervention~\cite{sun2019completely,sun2019evolving}. ENAS algorithms have accounted for the majority of existing NAS algorithms as evidenced from a recent survey paper~\cite{elsken2019neural}.

During past years, ENAS algorithms have attracted great attention owing to their high robustness, exceptional performance, and full ability to automate DNN architectures design~\cite{he2021automl,al2019survey}. For example, the LargeEvo algorithm~\cite{real2017large} firstly used GA to automate the discovery of DNN architectures for image classification. Meanwhile, the Genetic-CNN algorithm~\cite{xie2017genetic} proposed a fixed-length binary-string encoding method within GA to represent each network architecture. Furthermore, the Hierarchical-Evo algorithm~\cite{liu2017hierarchical} combined a new hierarchical genetic representation scheme to achieve efficient NAS. The EvoCNN algorithm~\cite{sun2019evolving} firstly developed the variable-length encoding strategy to search for promising DNN architectures without requiring manual effort during the search stage. The CGP-CNN algorithm~\cite{suganuma2017genetic} automatically constructed CNN architectures for image classification based on Cartesian genetic programming~\cite{miller2008cartesian}, which is a type of GP~\cite{miller2008cartesian}. In addition, the AE-CNN algorithm~\cite{sun2019completely} and the CNN-GA algorithm~\cite{sun2020automatically} proposed to automatically evolve CNN architectures by using GA based on state-of-the-art CNN blocks~\cite{2016Deep,huang2017densely}. The NSGA-Net~\cite{lu2019nsga} algorithm achieved NAS by considering multiple conflicting targets using multi-objective EC algorithms~\cite{deb2000fast}. The Regularized-Evo algorithm~\cite{real2019regularized} introduced age attributes in a modification of tournament selection of GA, evolving an image classifier surpassing manual design for the first time.

Although ENAS researchers have drawn more attention to the community, there are still two challenges that need to be addressed. Specifically, the two challenges are the fair comparison issue and the efficient evaluation issue, as will be detailed below.

The fair comparison issue is widely recognized in ENAS. Making fair comparisons between different ENAS algorithms for ENAS developers is challenging, if not impossible. This has somehow hindered the development of ENAS because unfair comparisons may mislead the researchers and resulted in frustration. In addition, the researchers cannot exactly know how novel or competitive their algorithms are. Since ENAS is widely used in image classification tasks, taking the image classification task as an example, the classification accuracy (or classification error rate) and the computation budget of the searched architectures are the two most popular indicators used to evaluate the performance of existing ENAS algorithms~\cite{liu2020survey}. When an ENAS algorithm is proposed and its performance is investigated on the classification accuracy, the developer needs to collect the classification accuracy values from some chosen state-of-the-art ENAS algorithms. However, because most ENAS algorithms are not of open-source, reproducing them under the same condition for arriving at fair classification accuracy is a nontrivial matter~\cite{li2020random,yu2019evaluating}. In the most common practice, the reported classification accuracies of peer competitors for comparison are often extracted directly from their respective seminal papers. However, the experimental setups of these peer competitors are vastly different from each other, such as using different data preprocessing~\cite{zhang2017mixup,devries2017improved,ghiasi2018dropblock}, different optimizers~\cite{szegedy2016rethinking,goyal2017accurate}, different learning rates, different batch sizes, and different training epochs, which are all deciding factors of the classification accuracy resulted by these ENAS algorithms. In this regard, the comparisons in terms of classification accuracy are clearly biased. 
On the other hand, the common practice for comparing with computation budget of ENAS algorithms is to measure by ``GPU days'' (i.e., the number of GPUs used $\times$ the days elapsed).
Due to the different types of GPUs used by different ENAS developers, it is difficult to directly compare the computation budget of different ENAS algorithms by directly citing from their seminal papers. As a result, the comparison in terms of the collected ``GPU days'' is also unfair. 
On the other hand, the number of function evaluations (i.e., the generation number $\times$ the population size) is a key metric to fairly compare the performance of EC methods. Based on the extensive observations from the seminal papers of existing ENAS algorithms, most of them used different population sizes and generation numbers that result in different function evaluation numbers, and this is even worse as the individuals in the population(s) are of variable-length instead of fixed-length. This again leads to unfair comparisons among ENAS algorithms.

In addition, the lack of efficient evaluation methods has also hindered the development of ENAS. The ENAS algorithms depend upon the heavy requirement of computational resources because the fitness evaluation of DNNs during the evolutionary search will consume a lot of time. Specifically, the fitness evaluation of a DNN is achieved by training the DNN on the target dataset via a training-from-scratch process, which is nevertheless time-consuming. For example, training a DNN on a common dataset such as CIFAR-10~\cite{krizhevsky2009learning} often consumes hours to days depending on the scale of the DNN. 
Moreover, since the EC methods are population-based~\cite{schmitt2001theory,banzhaf1998genetic,back1996evolutionary}, there are many individual DNNs to be evaluated during the search process of the ENAS, which greatly increases the computational overhead of ENAS.
For a common research environment, there are often multiple GPUs available in the lab. As a result, in-lab researchers often pursuit an acceleration of running ENAS algorithm by evaluating their fitness on multiple GPUs. 
There are usually two strategies for using multiple GPUs. 
One strategy is evaluating the fitness of individuals with the in-house distributed parallel methods from the deep learning libraries such as PyTorch~\cite{paszke2017automatic} and Tensorflow~\cite{abadi2016tensorflow}, and the other strategy is to use distributed NAS toolkits.
For the first strategy, these in-house distributed parallel methods use only multiple GPUs for the evaluation of a single DNN, which is not suitable for evaluating a large number of DNNs in ENAS. In this method, different GPUs need to communicate with each other to transfer computational parameters, which unavoidably increases the communication overhead and inadvertently increases the time for individual DNN training.
The second strategy is a more effective approach to speeding up the ENAS algorithm because users can use these distributed toolkits to evaluate multiple DNNs simultaneously during the fitness evaluation phase of ENAS. Under this strategy, one GPU is used to evaluate only one DNN and there is little communication overhead incurred between GPUs.
However, these existing distributed NAS toolkits, such as NNI are complex to configure and have high learning costs. Therefore, they are not necessarily friendly for in-lab users.

In this paper, we aim to develop a benchmarking platform of ENAS algorithms named BenchENAS, to address all the issues aforementioned. In summary, the contributions of the proposed BenchENAS are shown as follows:

\begin{table*}[htbp]
\caption{The 9 ENAS Algorithms included in the version 1.0 of BenchENAS}
\begin{center}
\small
\begin{tabular}{c|c|c}
\hline
\textbf{Algorithm}                  & \textbf{Year} & \textbf{Paper}                                                                                                     \\ \hline
LargeEvo~\cite{real2017large}                            & 2017          & Large-Scale Evolution of Image Classifiers                                                                         \\
CGP-CNN~\cite{suganuma2017genetic}                             & 2017          & A Genetic Programming Approach to Designing Convolutional Neural Network Architecture                              \\
Genetic CNN~\cite{xie2017genetic}                     & 2017          & Genetic CNN                                                                                                        \\
HierarchicalEvo~\cite{liu2017hierarchical}                 & 2018          & Hierarchical Representations for Efficient Architecture Search                                                     \\
AE-CNN~\cite{sun2019completely}                              & 2019          & Completely automated CNN architecture design based on blocks                                                       \\
EvoCNN~\cite{sun2019evolving}                              & 2019          & Evolving deep convolutional neural networks for image classification                                               \\
NSGA-Net~\cite{lu2019nsga}                            & 2019          & NSGA-Net: Neural Architecture Search using Multi-Objective Genetic Algorithm                                       \\
\multicolumn{1}{l|}{RegularizedEvo~\cite{real2019regularized}} & 2019          & \multicolumn{1}{l}{Regularized Evolution for Image Classfier Architecture Search}                                  \\
\multicolumn{1}{l|}{ \quad CNN-GA~\cite{sun2020automatically}}         & 2020          & \multicolumn{1}{l}{Automatically Designing CNN Architectures Using the Genetic Algorithm for Image Classification} \\ \hline
\end{tabular}
\label{tab1}
\end{center}
\end{table*}

\begin{itemize}
\item Nine representative state-of-the-art ENAS algorithms, popular data processing techniques for three widely used benchmark datasets, as well as configurable trainer settings such as learning rate policy, optimizers, batch size, and training epochs, have been implemented into the proposed BenchENAS platform. To this end, the researchers can illustrate the innovativeness of their proposed algorithms by making fair comparisons with the state-of-the-art ENAS algorithms. Furthermore, these algorithms cover fixed-length encoding strategies and variable-length encoding strategies, and also single-objective optimization algorithms and multi-objective optimization algorithms as shown in Table 1. We believe the proposed BenchENAS platform can meet the needs of most researchers.

\item BenchENAS has good usability and easy extensibility. BenchENAS is easy to use for four main reasons. Firstly, BenchENAS is implemented in python using very few third-party libraries for easy installation. Secondly, all the ENAS algorithms in BenchENAS can be easily configured with different data settings and different trainer settings. Thirdly, BenchENAS designs a downtime restart strategy to reboot the platform in the event of an unexpected stop to improve the stability and robustness.
Furthermore, BenchENAS is fully open-sourced and is promised to be free for research use. 
BenchENAS is easy to extend due to its highly modular design. Users can easily implement their own ENAS algorithms within BenchENAS. It is also easy for users to extend dataset settings as they become available such as benchmark datasets, data processing techniques, and trainer settings (e.g. learning rate policy, optimizers), etc. 

\item An efficient parallel component and a cache component are designed to accelerate the fitness evaluation phase in BenchENAS. The parallel component is specifically designed for ENAS algorithms for conveniently performing parallel training of DNNs by in-lab users, which is based on the parallel training mechanism of existing deep learning libraries and can be jointly used to collectively speed up the running of the corresponding ENAS algorithm. This platform mimics the use of manual GPU assignments in the lab environment, and can be used without any tedious requirements or setup from existing peer platforms. In addition, the designed method can also adaptively detect the available GPU in the environment and flexibly assign the free GPUs to the algorithms. The cache component is used to record the fitness values for each architecture and to reuse the fitness values in the cache when an individual of the same architecture appears.

\item Based on the designed BenchENAS platform, we have performed fair comparisons of the implemented ENAS algorithms on BenchENAS with popular settings. These experimental results can be used by researchers as benchmark data for future studies so that they will not need to rerun these algorithms for their comparison. The experimental results show that the comparison results within BenchENAS are appreciably different from those of the respective original papers, implying that unfair comparisons do exist. This in turn justifies the necessity of the proposed BenchENAS platform.
\end{itemize}

The rest of this paper is organized as follows. The related works of BenchENAS are reviewed in Section II. The details of BenchENAS are presented in Section III. Section IV illustrates the extensibility and usability of BenchENAS. Section V introduces the experimental result of BenchENAS. Finally, the conclusion and future work are given in Section VI.

\section{Related Works}
In this section, the related works of BenchENAS are presented. Specifically, the  background of ENAS algorithms is briefly documented in Subsection II-A, and then existing fair comparison approaches are introduced in Subsection II-B, and finally, popular efficient evaluation methods are shown in Subsection II-C.

\subsection{Background of ENAS}
NAS aims to automatically generate well-performing DNN architectures from a predefined search space using a well-designed search strategy. Mathematically, the NAS is generally considered as an optimization problem which is formulated by Equation (1):
\begin{equation}
\left\{\begin{aligned}
\arg \text{max}_{A} &=\mathcal{P}\left(A, \mathcal{D}_{\text {train }}, \mathcal{D}_{\text {valid }}\right) \\
&\text { s.t. } \quad A \in \mathcal{A}
\end{aligned}\right.
\end{equation}
where ${\mathcal{A}}$ denotes the search space of the neural architectures, $\mathcal{P}$(·) measures the performance of the architecture $A$ on the validation dataset $\mathcal{D}_{\text {valid }}$ after being trained on the training dataset $\mathcal{D}_{\text {train }}$. 
Essentially, NAS is a complex optimization problem experiencing multiple challenges, such as complex constraints, discrete representations, bi-level structures, computationally expensive characteristics and multiple conflicting objectives~\cite{sun2019completely}.

As introduced in Section I, ENAS is a subcategory of NAS. The main steps of an ENAS algorithm followed by the standard flowchart of an EC method are shown as follows.

\begin{enumerate}[step 1]
\item Initialize a population of individuals representing different DNN architectures within the predefined search space.
\item Evaluate the fitness of each DNN architecture. 
\item Select the parent solutions from the population based on the fitness values.
\item Generate offspring using evolutionary operators.
\item Go to step 6 when the criterion is satisfied; otherwise, go to step 2.
\item Terminate the evolutionary process and output the DNN with the best fitness value.
\end{enumerate}

As can be seen above, the ENAS algorithm follows the standard flowchart of an EC method~\cite{back1997handbook}. Specifically, step 2 shows the fitness evaluation phase of the ENAS algorithms. 
For the majority of ENAS algorithms~\cite{sun2019completely, sun2019evolving, real2019regularized,sun2020automatically}, the fitness value of each individual in the population will be estimated during the phase of fitness evaluation. In ENAS, the fitness value generally represents the performance of DNNs such as the accuracy on image classification tasks. 
In step 3, the mating selection operator acts as a natural selection increasing the quality of individuals because the individuals with superior performance are chosen as parent solutions to generate more competitive individuals, while ill-fitted individuals are eliminated. There are a lot of strategies in how to select an individual, such as random selection, tournament selection~\cite{liu2017hierarchical}, and the roulette wheel selection~\cite{de1975analysis}. 
In step 4, the evolutionary operators mainly include crossover and mutation. These operators create necessary diversity as well as novelty. Crossover recombines the evolutionary information of two parents to generate new offspring. Mutation, on the other hand, randomly changes the encoding of the DNN with a certain probability. 

Generally, the earliest work of ENAS is viewed as the LargeEvo algorithm. Since the success of LargeEvo, ENAS has gained great attention. More and more ENAS algorithms are proposed recently. Vast neural architectures automatically designed by these ENAS algorithms have surpassed manual designs in many tasks~\cite{real2019regularized,passricha2019pso}. Despite the positive results of the existing ENAS algorithms, there are still some challenges and issues which need to be addressed, such as the fair comparison issue and the efficient evaluation mentioned above.

\subsection{Fair Comparison Methods}
As have introduced in Section-I, the fair comparison issue may mislead researchers. More and more researchers are noticing the existence of the fair comparison issue in the NAS field and proposing some solutions. Representative works mainly include NAS-Bench-101~\cite{ying2019bench} and NAS-Bench-201~\cite{dong2020bench}. 

Specifically, these works enable fair comparisons by providing a benchmark architecture dataset for researchers. 
NAS-Bench-101 provides the first public architecture dataset for NAS Research. It constructed a search space, exploiting graph isomorphisms to identify 423K unique convolutional architectures. Then, they trained and evaluated all of these architectures three times on CIFAR-10 and compiled the results into a large dataset of over 5 million trained architectures. Each architecture can query the corresponding metrics, including test accuracy, training time, etc., directly in the dataset without the large-scale computation.
NAS-Bench-201 is proposed recently and is based on cell-based encoding space. Compared with NAS-Bench-101, which was only tested on CIFAR-10, this dataset collects the test accuracy on three different image classification datasets (i.e., CIFAR-10, CIFAR-100~\cite{krizhevsky2009learning}, and ImageNet-16-120~\cite{chrabaszcz2017downsampled}). However, the encoding space is relatively small and only contains 15.6K architectures.
Nevertheless, experiments with different ENAS methods on these benchmark datasets can obtain fair comparisons between different ENAS algorithms and it will not take too much time. 

Unfortunately, those works have many limitations in practical applications. Firstly, NAS-Bench-101 and NAS-Bench-201 are applicable to only a few datasets. Secondly, the search space of these works is fixed with a limited number of nodes and edge types in a cell. 
These works are only based on the cell-based encoding space and do not apply to methods based on other coding spaces.

\subsection{Efficient Evaluation Methods}
As have introduced above, the ENAS algorithms usually consume a lot of computational resources and time. In the context of labs that typically have multiple GPUs, in-lab users usually evaluate DNNs by two efficient evaluation methods. Specifically, these two methods are using in-house distributed parallel methods in deep learning libraries and using distributed NAS toolkits.

The in-house distributed parallel methods in deep learning libraries can be divided into two different categories: data parallelism and model parallelism.
In the data parallelism, the dataset is split into multiple sub-datasets. Every node hosts a copy of the DNN and trains the DNN on a subset of the dataset as shown in Fig. 1. 
Then, the values of the parameters are sent to the parameter server. After collecting all the parameters, they are averaged. This classical realization of the data parallelism is named Synchronous Stochastic Gradient Descent (SSGD)~\cite{povey2014parallel, shi2019distributed}. This method is not efficient because the nodes are forced to wait for the slowest one at each iteration.
Another method is called asynchronous SGD (ASGD)~\cite{zhang2015staleness, lian2018asynchronous}.
This method improves on SSGD by sending outdated parameters out of the network.
However, this method leads to a gradient staleness problem, which may result in slow convergence speed or even non-convergence of the model.
In the model parallelism, each node hosts a partition of the DNN, and the dataset needs to be copied to all nodes as shown in Fig. 2. The communication happens between computational nodes when the input of a node is from the output of the other computational node. The model parallelism is not commonly used because the communication expense is much higher.
In addition, these distributed parallel methods are mainly targeted at a single large-scale DNN while a large number of DNNs needed to be evaluated simultaneously in ENAS. Specifically, assuming that there are $N$ GPUs in a lab, these methods only support accelerating the training of a single DNN with these $N$ GPUs. In this case, however, using data parallelism can lead to synchronization overhead or cause gradient staleness problems. Using model parallelism can lead to unnecessary communication overhead. In the context of a large number of DNNs to be trained simultaneously, it is more efficient to train $N$ individuals simultaneously with $N$ GPUs to avoid unnecessary overhead. 

\begin{figure}[htbp]
\centerline{\includegraphics[width=0.35\textwidth]{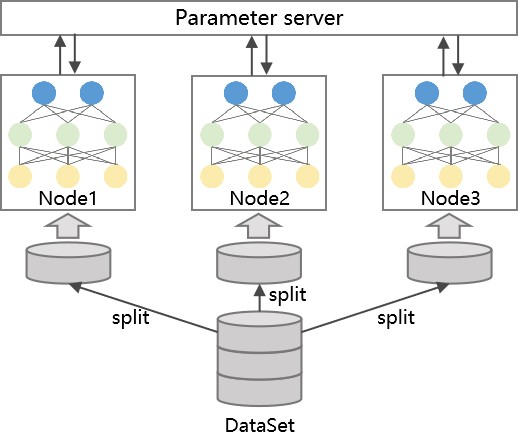}}
\caption{The architecture of the data parallelism.}
\end{figure}

\begin{figure}[htbp]
\centerline{\includegraphics[width=0.48\textwidth]{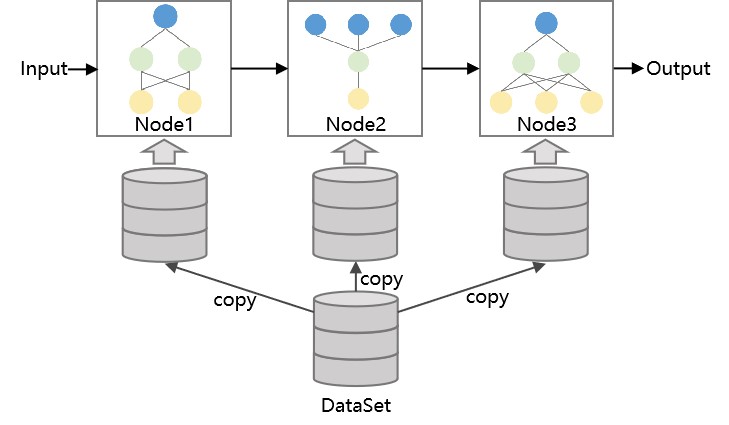}}
\caption{The architecture of the model parallelism.}
\end{figure}

In fact, some distributed NAS toolkits exist that support users to train $N$ DNNs simultaneously with $N$ GPUs. Representative works mainly include NNI\footnote{https://github.com/microsoft/nni} developed by Microsoft and Katib\footnote{https://github.com/kubeflow/katib/trial} at Google. Specifically, NNI has a built-in lightweight distributed training platform and trains multiple DNNs in parallel. It also supports the configuration of OpenPAI \footnote{https://github.com/microsoft/pai}, Kubernetes~\cite{bernstein2014containers} and some other distributed scheduling platforms for distributed training of multiple DNNs. Katib only supports the configuration of Kubernetes. However, those toolkits are not friendly to in-lab users. Firstly, they are not easy to extend. Commonly, the GPUs in the lab are distributed across multiple machines. To use the GPUs on multiple machines, those toolkits need to be installed on every machine. Secondly, these toolkits are more difficult to learn.
In order to use some distributed functions of those toolkits, users may need to learn to use existing distributed platforms (Kubernetes, OpenPAI), such as deploying the platform, configuring NVIDIA plug-ins, setting up storage servers, and many other operations.

In summary, the available fair comparison methods and the efficient evaluation methods bear some serious shortcomings. The current fair comparison methods are very limited and do not apply to all the main types of ENAS. In the current efficient evaluation methods, the built-in distributed parallel methods do not take into account the population characteristics of ENAS and instead increases the overhead. The distributed NAS toolkits are complicated to configure and use, and are not suitable for in-lab users.

\section{The proposed BenchENAS}
\subsection{Overview}
\begin{figure*}[htbp]
\centerline{\includegraphics[width=1.0\textwidth]{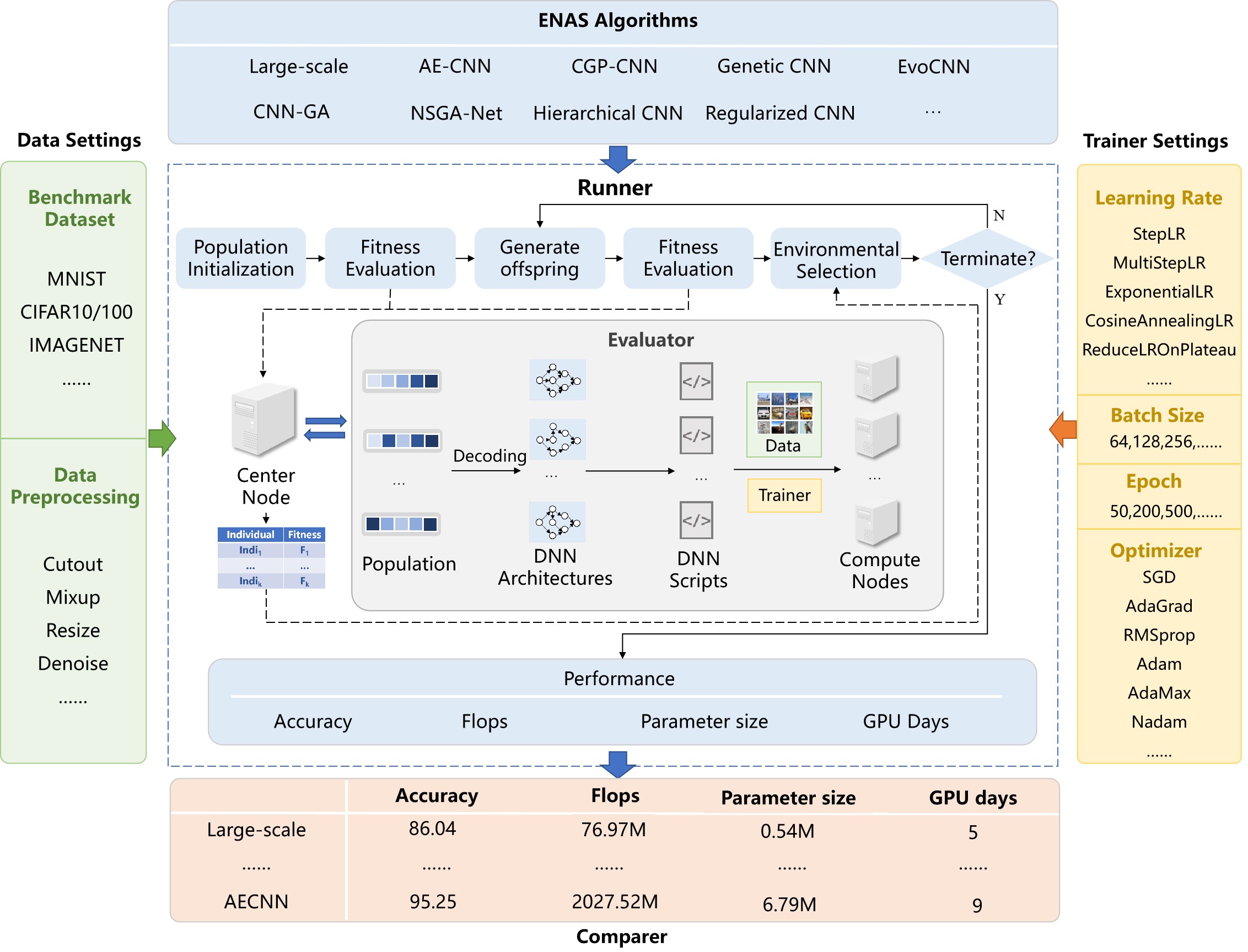}}
\caption{The overview of BenchENAS.}
\label{Fig. 2}
\end{figure*}
Fig. 3 shows the overview of BenchENAS. BenchENAS is composed of five parts. They are $ENAS\_algorithms$, $data\_settings$, $trainer\_settings$, $runner$, and $comparer$. 
After the user has selected the ENAS algorithm to run in the $ENAS\_algorithms$ part, the data and the trainer for the training of DNNs are configured through the $data\_settings$ part and the $trainer\_settings$ part. The selected ENAS algorithm is then run through the runner part and the results are output to the $comparer$ part. By running multiple ENAS algorithms with the same data settings and the same trainer settings in the $runner$ part, users can get relatively fair comparison results.

Specifically, the $ENAS\_algorithms$ part includes the ENAS algorithms implemented by BenchENAS. Users can choose the algorithm they want to run or implement their own code.
The $data\_settings$ part includes settings for the benchmark dataset and data preprocessing. 
Users can choose MNIST~\cite{lecun1998gradient}, CIFAR-10, CIFAR-100, or ImageNet as benchmark dataset for comparison. All of these datasets are widely used in the field of image classification. 
Users can also choose data preprocessing methods such as cutout~\cite{devries2017improved}, mixup~\cite{zhang2017mixup}, resize, and denoise. An option is also made available to incorporate any new utilities for data preprocessing. These data preprocessing methods can be used to improve the performance of DNNs. Specifically, cutout and mixup are both commonly used data augmentation methods to improve the generalization of neural network architectures. Cutout masks out random sections of input images during training. Mixup trains a neural network on convex combinations of pairs of examples and their labels, regularizing the neural network to favor simple linear behavior in-between training examples.
For the resize method, rescaling and cropping can be used to resize data.
Denoising removes useless information from the data to improve the performance of neural architectures.
The $trainer\_settings$ part includes learning rate settings, batch size settings, epoch settings, and optimizer settings. Users can set the learning rate by StepLR, MultiStepLR, ExponentialLR, CosineAnnealingLR~\cite{loshchilov2016sgdr}, and ReduceLROnPlateau learning rate strategies. These strategies are all very practical learning rate adjustment strategies. StepLR is a strategy to adjust the learning rate at equal intervals. MultiStepLR is a strategy to adjust the learning rate at set intervals. ExponentialLR is a strategy to adjust the learning rate by an exponential decay. CosineAnnealingLR is a strategy that takes the cosine function as the period and resets the learning rate at each period. ReduceLROnPlateau is a strategy that adjusts the learning rate when a metric is no longer changing.
The runner part is used to run the ENAS algorithm. As shown in Fig. 3, the center node launches and runs the entire ENAS algorithm. When the algorithm proceeds to the fitness evaluation phase, the runner uses a well-designed evaluator to obtain the fitness of each individual in the population.
In the evaluator, the center node firstly decodes the individuals in the population into DNN architectures. Secondly, the center node generates DNN scripts from the DNN architectures. Thirdly, the center node distributes the DNN scripts to the compute nodes. Each GPU in these compute nodes runs a DNN script with the data and the trainer to obtain the fitness of the DNN. Finally, each compute node sends the fitness value to the center node.
The comparer part includes the performance indicators such as the accuracy, flops, parameter size, and GPU days of different ENAS algorithms. Users can do fair comparisons between different ENAS algorithms by those performance indicators.
The core of BenchENAS is the runner part.
The details of the implementation of the runner are given in Section III-B.

\subsection{Runner}

\begin{algorithm}
\caption{The runner of BenchENAS}\label{algorithm}
\KwIn{Parameters required for the algorithm, the population size, the maximal generation number, the image dataset for classification, is_running.}
\KwOut{The discovered best DNN architecture.}
\eIf {is_running == 0} {
$t\leftarrow $\ 0$ $\; 
$P_{t}\leftarrow $\ Initialize a population$ $\; 
$is\_running \leftarrow $\ 1$ $\;
}{
$t\leftarrow $\ Get the newest generation number$ $\; 
$P_{t}\leftarrow $\ Load the population$ $\;  
}
{
Evaluate the fitness of each individual in \emph{P$_{t}$} \textbf{using the evaluator}  \; 
}
{Save the population \emph{P$_{t}$} \;}
{
\While{t \textless the maximal generation number}{
$Q_{t}\leftarrow $\ Generate offspring$ $\; 
Evaluate the fitness of each individual in \emph{P$_{t}$} \textbf{using the evaluator}  \; 
$P_{t+1}\leftarrow $Environmental selection from$ $\ $ \emph{P$_{t}$} \cup \emph{Q$_{t}$} $$ $\; 
Save the population \emph{P$_{t+1}$} \;
$t\leftarrow $\ \emph{t} + 1$ $\;
}
}
{$is\_running \leftarrow $\ 0$ $\;}
$\textbf{Return}$ $P_{t}.$
\end{algorithm}

Runner is used to run the ENAS algorithm to get the results.
The ENAS algorithm maintains a high degree of uniformity, and the implementation process is basically the same except for some minor differences in details. 
To increase the stability and robustness of BenchENAS, we design a log-based downtime restart strategy to allow BenchENAS to use logs to restart when it is unexpectedly shut down. Next, we will describe in detail the operation process of the runner with the downtime restart strategy.

The pseudocode of the runner with the downtime restart strategy is shown in Algorithm 1. Specifically, for the downtime restart strategy, we add a parameter called \emph{is_running} to the runner. ``\emph{is_running} == 0'' means that the runner has not run or finished running last time. On the other hand, ``\emph{is_running} == 1'' implies that the operation of the last runner was interrupted. When the runner starts running, if \emph{is_running} is equal to 0, the population \emph{P$_{0}$} is initialized and \emph{is_running} is set to 1 (Lines 1-4). Otherwise, the latest population counter \emph{t} is gotten, and the population \emph{P$_{t}$} is loaded according to the acquired population number (Lines 5-8). Then, the fitness of each individual, which encodes a particular architecture of the DNN, is evaluated on the given dataset by the evaluator (Line 9). The population is saved (Line 10).  During evolution, the parent individuals are selected based on the fitness, and then a new offspring is generated by the evolutionary operators, including the crossover and mutation operators (Line 12). After that, the fitness of each individual is evaluated using the evaluator (Line 13). The evaluator is designed by BenchENAS to save evaluation time and will be explained in detail in the next section. Then, a population of individuals surviving into the next generation is selected by the environmental selection from the current population (Line 14). Next, the population is saved (Line 15). Finally, the counter is increased by one, and the evolution continues until the counter exceeds the predefined maximal generation.
When the evolution is finished, the parameter \emph{is_running} is set to 0 (Line 18). 

Next, we will describe how to load the population information (Line 7). From Algorithm 1, we can find that the population information is saved after the fitness evaluation phase (Lines 10, 15). Specifically, BenchENAS saves the population \emph{P$_{t}$} as a file named \emph{begin_t.txt}. The file \emph{begin_t.txt} contains the name, the encoding information, the identifier, and the fitness value of each individual in the population. When \emph{is_running} is not 0, the latest written population file will be loaded and the information of each individual in the population will be obtained to rebuild the population \emph{P$_{t}$}. The downtime restart strategy enables log-based downtime restart. Doing so gives BenchENAS the advantages of persistence and stability. When a power outage or downtime is caused by the wrong operations, the saved log files are used to restore the platform to the working state before the downtime, thus avoiding wasting computational resources by starting training from scratch.

In the next sub-section, we will detail the implementation of the evaluator which is well-designed to save the computational resources and evaluation time.

\subsection{Evaluator}
As mentioned above, because training DNNs is very time-consuming, varying from several hours to even several months depending on the particular architecture, the evaluator is designed to speed up the fitness evaluation phase in BenchENAS.
In the evaluator, we reduce the evaluation time and the computational resources by the cache component and the parallel component. 
Specifically, the cache component is used to store the fitness of every DNN evaluated. It works by reusing the fitness of the DNNs that have previously appeared in the population to save time.
The parallel component works by evaluating multiple individuals on multiple GPUs.
Next, we will describe in detail how the evaluator uses these two components.

The detail of the evaluator is shown in Algorithm 2, which includes the specifics of the cache component (Lines 4-15). 
Briefly, given the population \emph{P$_{t}$} containing all the individuals for evaluating the fitness, the evaluator evaluates each individual of \emph{P$_{t}$} in the same manner, and finally returns \emph{P$_{t}$} containing the individuals whose fitness have been evaluated.
Specifically, for each individual in the population, the evaluator firstly decodes the individual into DNN and generates the python script of the DNN (Lines 2-3). The cache component method is achieved in Lines 4-15. Specifically, if the cache does not exist, the evaluator will create an empty global cache system (denoted as $Cache$), storing the fitness of the individuals with unseen architectures (Lines 4-7). Then, if the individual is found in $Cache$, its fitness is directly derived from $Cache$ (Lines 8-10). Otherwise, the individual is asynchronously evaluated using the parallel component to obtain the fitness of the individual (Lines 12-13). The identifier and fitness value of the individual will be stored to $Cache$ (Line 14). For the cache component, querying an individual from $Cache$ is based on the individual's identifier. Theoretically, arbitrary identifiers can be used as long as they can distinguish individuals encoding different architectures. In BenchENAS, the 224-hash code~\cite{housley2004224}, which has been implemented by most programming languages in terms of the encoded architecture is used as the corresponding identifier. For the parallel component, the individual is asynchronously placed on an available GPU, which implies that multiple individuals can be evaluated simultaneously on multiple GPUs.

The cache component is designed to speed up the fitness evaluation phase of the ENAS algorithms, which is mainly based on the following two considerations.
Firstly, individuals who survive to the next generation without changes in neural network architecture do not need to be re-evaluated.
Secondly, the evolutionary operators such as crossover and mutation may generate individuals that have been evaluated before. 
In such a context, the cache component can be used to save time and improve evolutionary efficiency. 
In general, we should be concerned about the size of the cache component and discuss the conflicting collision problem resulted from the duplicate keys.
However, in BenchENAS, it is not necessary a concerning issue. 
Firstly, the cache component is similar to a map data structure.  
Each of these records in the cache component is a string containing the identifier of a DNN and the corresponding fitness value.
For example, a record such as ``id = 90.50'' denotes that the identifier of the DNN is ``id'' and its fitness value is ``90.50''.
Secondly, as we mentioned in the last paragraph, the identifier is calculated by the 224-hash code that can generate 2$^{224}$ different identifiers.
Commonly, the ENAS algorithms only evaluate thousands of individuals. Obviously, the issue will hardly happen.
Thirdly, the 224-hash code implementation used by BenchENAS will generate the identifier with the length of 32, the symbol ``='' with the length of 1, and the fitness value with the length of 4. 
Thus, each record in the cache component is a string with the length of 37, occupying 37 bytes with the UTF-8 file encoding. 
Obviously, the cache component will occupy less than 1MB of disk space, even though there are tens of thousands of records. 
Therefore, we do not need to consider the size of the cache component.

\begin{algorithm}
\caption{Evaluator}\label{algorithm}
\KwIn{The population $P_{t}$ of the individuals to be evaluated.}
\KwOut{The population $P_{t}$ of the individuals with their fitness values.}
\ForEach {individual in $P_{t}$}
{
Decode the $individual$ to DNN\;
Generate the python script for the DNN\;
\If {$Cache$ does not exist} {$Cache\leftarrow \emptyset $\; Set \emph{Cache} to a global variable\;}
\eIf{the identifier of individual in Cache}{$v\leftarrow $\ Query the fitness by \emph{identifier} from \emph{Cache}$ $\; Set \emph{v} to \emph{individual}\;}{
{$\emph{v} \leftarrow $\ evaluate \emph{individual} \textbf{by the parallel component}$ $\;}
Set \emph{v} to \emph{individual}\;
{$\emph{Cache} \leftarrow $\ the identifier of individual and the fitness$ $\;}
}
}

$\textbf{Return}$ $P_{t}.$
\end{algorithm}

The parallel component is a parallel computing platform based on GPUs. As shown in Algorithm 2, when the identifier of the individual is not in $Cache$, the evaluator will evaluate the individual by the parallel component. This component is designed for the ENAS algorithm because the EC algorithms are population-based. In this component, a GPU is used to evaluate only one DNN. Assuming a total of $N$ GPUs for all compute nodes, $N$ individuals are evaluated simultaneously during the fitness evaluation phase of ENAS. Specifically, the center node gets GPU usage from a SQL database. When the ENAS algorithm begins to run, the SQL database including the state (i.e. usage) of each GPU is created. The center node remotely queries the GPU usage of all compute nodes in parallel at a regular intervals. 
When the fitness of an individual is to be obtained, the evaluator will use the parallel component to evaluate the individual as Algorithm 3 shows. Specifically, while there is an available GPU by querying the SQL database, the center node firstly gets the compute node, says \emph{node$_{j}$}, and the identifier of \emph{GPU$_{k}$}. Secondly, the center node set the usage of \emph{GPU$_{k}$} to busy in the SQL database. Then, the center node sends the DNN script to \emph{node$_{j}$} and remotely commands \emph{node$_{j}$} to train the DNN script with \emph{GPU$_{k}$} (Lines 2-7). Finally, when the training of the DNN script is completed, the fitness value of the DNN script will be obtained. And the status of the \emph{GPU$_{k}$} is updated again in the SQL database. 

Next, the reasons for designing such a parallel component are given. 
Since training DNNs can take a lot of time, in-lab users often run the ENAS algorithm with multiple GPUs to speed up the evaluation. 
As discussed in related works, current methods of running ENAS with multiple GPUs such as the in-house distributed parallel methods and the distributed NAS toolkits are not suitable for in-lab users.
This motivates us to design such a parallel component.
The parallel component is designed using parallel computing techniques. In parallel computing, large problems can often be divided into multiple independent subproblems, which can then be solved at the same time. By parallel performing these subproblems in different computational platforms, the total processing time of the entire problem is consequently shortened. In the fitness evaluation phase of ENAS, multiple individuals are waiting to be evaluated at the same time due to the population-based characteristics. Furthermore, the fitness evaluation of each individual is independent, which just satisfies the scene of using the parallel computing techniques. As a result, we design the parallel component.
For the parallel component, the individual is placed in parallel on an available GPU, which implies that we do not need to wait for the fitness evaluation of the next individual until the fitness evaluation of the current one finishes, but place the next individual on an available GPU immediately. In doing so, BenchENAS significantly reduces the time consumed by the fitness evaluation.

\begin{algorithm}
\caption{The parallel component}\label{algorithm}
\KwIn{The individual $indi_{i}$ to be evaluated.}
\KwOut{The fitness values of $indi_{i}$.}
\While{there is an available GPU in the SQL database}{
$node_{j}\leftarrow $\ get the compute node where the GPU located$ $\; 
$GPU_{k}\leftarrow $\ get the GPU id$ $\; 
Set $GPU_{k}$ to busy in the SQL database\; 
Send the DNN script of $indi_{i}$ to $node_{j}$\; 
$Fitness_{i}\leftarrow $\ Remote command $node_{j}$ train the DNN script with $GPU_{k}$$ $\; 
Set $GPU_{k}$ to the idle state in the SQL database\; 
}
$\textbf{Return}$ $Fitness_{i}.$
\end{algorithm}

\begin{figure}[htbp]
\centerline{\includegraphics[width=0.4\textwidth]{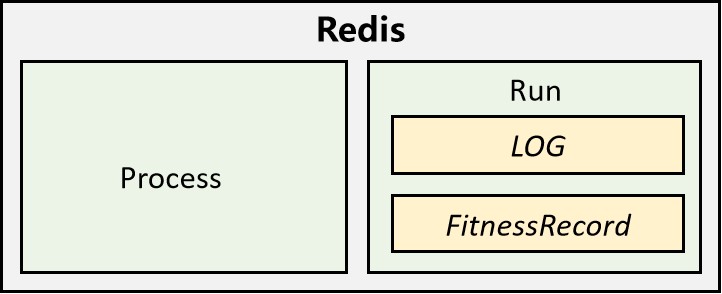}}
\caption{The Category of Redis database.}
\end{figure}

During the operation process of the parallel component, the information generated by all compute nodes needs to be exported to the center node through the information integration module. As mentioned above, the parallel component divides the problem of evaluating $N$ individuals in the fitness evaluation phase of ENAS into $N$ subproblems. Each subproblem is solved by a sub-process that trains a DNN on one GPU of a random compute node.
Each sub-processes will generate lots of information which can be firstly divided into two main categories as shown in Fig. 4. 

One category is $Process$ information, and the other category is $RUN$ information.
The Redis database is used to transfer these two types of information between the center node and compute nodes. 
Specifically, $Process$ information mainly contains the IP address of the compute node where the sub-process is located and the pid of the sub-process. Before the sub-process starts training the DNN, the sub-process inserts the process information into the Redis database. When training is complete, the sub-process will remove its own process information from the Redis database. This allows the user to keep track of the processes running on each compute node. In addition, when the user interrupts BenchENAS, BenchENAS will kill each sub-process to prevent it from continuing to occupy computational resources. 
$RUN$ information mainly contains the $LOG$ record output by the sub-process during training and the $FitnessRecord$ record obtained when the training is completed. The $LOG$ record is the log of the output of the sub-process during training, such as epoch and the corresponding accuracy.  
The $FitnessRecord$ record is mainly used to output the fitness values obtained from training to two types of files: Cache and Result. The Cache file is the implementation of the cache component introduced before, and the sub-process will output the individual identifier and its fitness value to the cache when it obtains the individual’s fitness value. BenchENAS uses this cache file as the cache component, reducing the time consumed by the evaluation. The Result file is a hash table that stores the fitness value of each individual. When the sub-process obtains the fitness value, the sub-process outputs the individual name and its fitness value to the Result file. It is convenient for users to view the fitness values of each individual at any time through this file.
When there are $RUN$ records that need to be exported to the center node, the sub-process inserts the information into the Redis database. 
There is a listener process that keeps looking at the Redis database, and when records exist, the process performs the appropriate operation based on the record type and deletes the record at the end of the operation to avoid duplicate reads. For example, if a $FitnessRecord$ record that needs to be stored in the cache is inserted into the Redis database, the listener process will read the record, insert the fitness value into the cache file, and delete the record.

The reason that we choose the Redis database is that the speed of information output of each sub-process is very fast during training. The Redis database is a high-performance key-value database. Redis can read data 110,000 times per second and write data 81,000 times per second. Therefore, Redis database meets the need for the information output. In this way, we consolidate all the information that we want to output to the center node during the fitness evaluation phase. This has two advantages. One is that users can view the log information and population information in the process of operation at any time, and analyze the results of operation. Users do not need to wait for the algorithm to finish running before analyzing the running process; Second, it is convenient for users to compare the difference between the running results of different algorithms.

\section{Usability And Extensibility}
In this section, we will explain the merit of BenchENAS in terms of its good usability and easy extensibility.

\begin{table*}[]
\caption{The global.ini for BenchENAS}
\normalsize
\begin{center}
\begin{tabular}{c|c|c|c}
\hline
\textbf{Section}           & \textbf{Key}   & \textbf{Type}     & \textbf{Description}         \\ \hline
\multirow{4}{*}{algorithm} & name           & string            & Folder name of the algorithm output    \\
                           & run\_algorithm & positive interger & Name of the algorithm used \\
                           & max\_gen       & positive interger & Max generation size              \\
                           & pop\_size      & positive interger & Population size              \\ \hline
\end{tabular}
\end{center}
\end{table*}

\begin{table*}[]
\caption{The train.ini for BenchENAS}
\normalsize
\begin{center}
\begin{tabular}{c|c|c|c}
\hline
\textbf{Section}           & \textbf{Key}      & \textbf{Type}     & \textbf{Description}       \\ \hline
\multirow{3}{*}{optimizer} & \_optimizer\_name & string            & Name of the optimizer used \\
                           & \_batch\_size     & positive interger & Batch size                 \\
                           & \_total\_epoch    & positive interger & Number of epoch            \\ \hline
\multirow{2}{*}{LearningRate}     & lr                & float             & Learning rate              \\
                           & lr\_strategy      & string            & Learning rate strategy     \\ \hline
dataset                    & \_name            & string            & Name of the dataset used   \\ \hline
\end{tabular}
\end{center}
\end{table*}

\begin{figure*}[htbp]
\centerline{\includegraphics[width=0.85\textwidth]{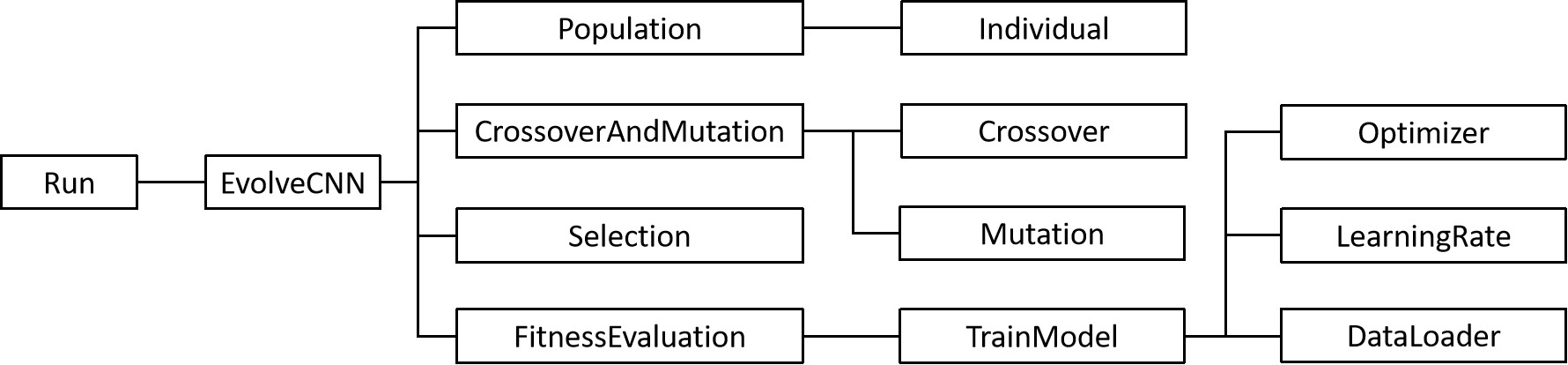}}
\caption{Relationship diagram between classes}
\end{figure*}

\subsection{Good Usability}
BenchENAS is easy to use. 
First of all, BenchENAS is easy to install. Specifically, BenchENAS is implemented in python and uses very few third-party libraries, except for the built-in libraries in python itself, which are torch, torchvision, paramiko and redis. 
Secondly, users only need to install BenchENAS on the center node, and on the compute node the user only needs to configure the computing environment for training DNNs.
Furthermore, BenchENAS is very easy to configure. Users can easily select the ENAS algorithm used, data settings and trainer settings through several $ini$ files. Specifically, the name of the ENAS algorithm used, the max generation and population size can be configured via a file named $global.ini$ as shown in Table II. 
The data settings and the trainer settings can be configured via a file named $train.ini$ as shown in Table III. 
When evaluating individuals, BenchENAS loads the optimizer, batch size, epoch, learning rate, and data for the evaluation by using the configuration in $train.ini$.
BenchENAS implements several optimizer classes such as Adam, RMSprop, SGD, several LearningRate classes such as CosineAnnealingLR, ExponentialLR, MultiStepLR and several dataloader classes for MNIST, CIFAR-10/CIFAR-100, ImageNet. 
Users select the optimizer to be used by filling in the optimizer class name in the $\_optimizer\_name$ in $train.ini$. 
Similarly, the learning rate strategy and the dateset are selected by the values in $train.ini$. 
In addition, the batch size, epoch and initial value of learning rate to be used can also be set via $train.ini$.
Furthermore, BenchENAS uses a log-based downtime restart strategy as mentioned above. When the downtime is caused by power failure or wrong operation, the saved information and log files are used to restore BenchENAS to the working state before the downtime, thus avoiding wasting computational resources by starting training from scratch again. This makes the computing platform with advantages of persistence and stability.
Finally, BenchENAS is completely open-sourced, and users can add their own algorithms. 
We will make the code publicly available upon request.

\subsection{Easy Extensibility}
BenchENAS is an open platform for the research of ENAS, hence it allows users to extend their own ENAS algorithms. 
BenchENAS is easy to expand in two ways. 
On one hand, users can easily write their own ENAS algorithms in BenchENAS with their defined evolutionary operators such as crossover, mutation, and selection.
On the other hand, users can easily add the contents of the data settings and trainer settings such as datasets, optimizers as shown in Fig. 3. 

The reason that users can easily write their own ENAS algorithms is that BenchENAS is highly modular. 
BenchENAS has a simple architecture, where it involves 13 classes.
Specifically, the relationships between these classes are illustrated in Fig. 5. The user starts BenchENAS via the class $Run$.
The class $Run$ is implemented by running the class $EvolveCNN$. 
The entire process of running the ENAS algorithm is defined in $EvolveCNN$.
The class $EvolveCNN$ consists of the class $Population$, the class $CrossoverAndMutation$, the class $Selection$, and the class $FitnessEvaluation$.
The user can change the entire flow of the ENAS algorithm by changing $EvolveCNN$, such as removing the crossover operation from the ENAS algorithm or changing the stopping condition.
One of the classes, $Population$, which is the class that defines the population, is implemented by the class $Individual$. Users can define their own $Individual$ to implement operations such as changing the encoding method and properties of individuals.
The class $CrossoverAndMutation$ is the class that implements crossover and mutation to generate new offspring, which includes the class $Crossover$ and the class $Mutation$. Users can define their own $Crossover$ and $Mutation$.
The class $Selection$ is used for environment selection, eliminating the ill-fit individuals and selecting the competitive ones.
The class $FitnessEvaluation$ gets the fitness value of each individual in a population by training each individual using the class $TrainModel$. In general, the class $FitnessEvaluation$ does not need to be changed because the process of evaluation should be the same for each individual in BenchENAS.
These classes are placed in folders with the corresponding names, and users can easily implement their own algorithms by changing these classes.

For data settings and trainer settings, the class $TrainModel$ is implemented by loading the class $Optimizer$, the class $Learningrate$ and the class $DataLoader$.
BenchENAS has provided several $Optimizer$, $Learningrate$ and $DataLoader$ classes.
Users can customize $Optimizer$, $Learningrate$ and $DataLoader$ classes to extend the optimizer settings and data settings. 
As for batch size, epoch, and initial value of learning rate, users can expand them by filling in new values in $train.ini$.

\section{Experiments and Analysis }
In this section, we run eight ENAS methods on BenchENAS, which can serve as baselines for future research. 
Since the original paper of EvoCNN did not perform experiments on the CIFAR-10 dataset, it does not participate in the comparison of ENAS algorithms.
Specifically, we evaluate some typical ENAS algorithms, including LargeEvo~\cite{real2017large}, Genetic CNN~\cite{xie2017genetic}, HierarchicalEvo~\cite{liu2017hierarchical}, CGP-CNN~\cite{suganuma2017genetic}, AE-CNN~\cite{sun2019completely}, NSGA-Net~\cite{lu2019nsga}, RegularizedEvo~\cite{real2019regularized}, CNN-GA~\cite{sun2020automatically}. Noting that the experiments of BenchENAS are performed on GPU cards with the same model of Nvidia GeForce GTX 2080 Ti.

To ensure a consistent evolution across ENAS algorithms, for each algorithm, the number of function evaluations is set at 400. We ensure that the other settings of the algorithm are consistent with those cited in their original papers, such as mutation probability, network length, convolution layer settings, etc. Due to the limitation of computing resources, we used the early stop policy in our experiments. We divide the algorithm into the evolution phase and the re-train phase. 
The evolutionary phase is the search process of ENAS, in which only the best performing DNN architecture needs to be found, so it is not necessary to train each DNN to its best performance. The approximate performance of each DNN can be observed by training epochs using the early stop policy.
In the evolution phase, we use 50 epochs to train the DNNs to reduce the evolution time. In the re-train phase, we retrain the best DNN selected by the algorithm to reach its best performance, and this phase uses 600 epochs. The specific settings will be explained in Section V-B.

\begin{table*}[]
\caption{Experimental results of each algorithm in the \textbf{original} paper on \textbf{CIFAR-10}}
\begin{center}
\normalsize
\begin{tabular}{c|c|c|c|c}
\hline
\textbf{Algorithm} & \textbf{Accuracy} & \textbf{\#Parameters} & \textbf{GPU days} & \textbf{Number of function evaluations} \\ \hline
LargeEvo~\cite{real2017large}           & 94.60             & 5.4M                  & 2750              & -                        \\
CGP-CNN~\cite{suganuma2017genetic}             & 94.02             & 1.68M                 & 27                & 1000                           \\
Genetic CNN~\cite{xie2017genetic}        & 92.90             & -                     & 17                & 1000                           \\
HierarchicalEvo~\cite{liu2017hierarchical} & 96.37             & -                     & 300               & 1400000                        \\
AE-CNN~\cite{sun2019completely}             & 95.7              & 2.0M                  & 27                & 400                            \\
NSGA-Net~\cite{lu2019nsga}           & 96.15             & 3.3M                  & 8                 & 1200                           \\
RegularizedEvo~\cite{real2019regularized}     & 96.60             & 2.6M                  & -                 & 20000                          \\
CNN-GA~\cite{sun2020automatically}             & 95.22             & 2.9M                  & 35                & 400                            \\ \hline
\end{tabular}
\end{center}
\end{table*}

\begin{table*}[]
\caption{Experimental results of each algorithm on \textbf{BenchENAS}}
\begin{center}
\normalsize
\begin{tabular}{c|c|c|c|c}
\hline
\textbf{Algorithm} & \textbf{Accuracy After 50 epoch} & \textbf{Re-train Accuracy} & \textbf{\#Parameters} & \textbf{GPU days} \\ \hline
LargeEvo           & 80.55                       & 86.04                      & 0.54M                 & 5                 \\
CGP-CNN            & 89.07                       & 94.24                      & 4.62M                 & 14                \\
Genetic CNN        & 85.36                       & 92.86                      & 0.28M                 & 3                 \\
HierarchicalEvo    & 85.59                       & 95.26                      & 47.13M                & 4                  \\
AE-CNN             & 89.04                       & 95.25                      & 6.79M                 & 9                 \\
NSGA-Net           & 85.12                       & 93.08                      & 1.08M                 & 4                 \\
RegularizedEvo     & 89.97                       & 94.42                      & 7.88M                 & 15                \\
CNN-GA             & 90.85                       & 94.67                      & 2.86M                 & 5 \\ \hline
\end{tabular}
\end{center}
\end{table*}

\subsection{Data Settings}
The CIFAR-10 benchmark dataset is chosen as the image classification task in the experiments. 
There are two reasons to choose it. 
Firstly, the dataset is challenging in terms of the image sizes, categories of classification, and noise as well as rotations in each image.
Secondly, it is widely used to measure the performance of ENAS algorithms, and most of the ENAS algorithms have publicly reported their classification accuracy on it. 
Specifically, the CIFAR-10 dataset is an image classification benchmark for recognizing ten classes of natural objects, such as airplanes and birds. It consists of 60,000 RGB images in the dimension of 32 × 32. In addition, there are 50,000 images and 10,000 images in the training set and the testing set, respectively. Each category has an equal number of images. 
In order to do a fair comparison, we employ the data preprocessing method that is often used in ENAS algorithms~\cite{2016Deep, huang2017densely, sun2020automatically}. 
Specifically, each direction of one image is padded by four zeros pixels, and then an image with the dimension of 32 × 32 is randomly cropped. Finally, a horizontal flip is randomly performed on the cropped image with a probability of 0.5. In this re-training phase, we use the additionally data preprocessing technique cutout. 

\subsection{Trainer Settings}
To assign the fitness to the candidate DNN architectures, we train the DNN by stochastic gradient descent (SGD) with a momentum of 0.9, a mini-batch size of 64, and a weight decay of 5.0×10$^{-4}$. The softmax cross-entropy loss is used as the loss function. We train each DNN for 50 epochs at an initial learning rate of 0.025.

After the evolution process, we re-train the best DNN architecture. In this re-training phase, we optimize the weights of the best architecture for 600 epochs with a different training procedure. We use SGD with a momentum of 0.9, a mini-batch size of 96, and a weight decay of 3.0×10$^{-4}$. We start with a learning rate of 0.025 and use the cosineAnnealing learning rate.

\subsection{Result And Analysis}
In our experiments, we mainly compare the classification accuracy, parameter size and GPU days for different ENAS algorithms. For the convenience of summarizing the comparison results, we use the name of the ENAS algorithms as the name of the discovered best DNN when comparing the classification accuracy, parameter size and GPU days between different ENAS algorithms.

The accuracy, parameter size, GPU days, and the number of function evaluations in the original paper are shown in Table VII. The accuracy after 50 epochs on the test dataset, re-train accuracy, parameter size, and GPU days of these ENAS algorithms in BenchENAS are shown in Table VIII. The number of function evaluations of these algorithms is 400. The symbol ``–'' implies there is no result publicly reported by the corresponding algorithm. 

We can find from the experimental results that, in terms of accuracy, the accuracy ranking of all algorithms derived from the original papers is: RegularizedEvo \textgreater HierarchicalEvo \textgreater NSGA-Net \textgreater AECNN \textgreater CNN-GA \textgreater LargeEvo \textgreater CGP-CNN \textgreater Genetic CNN with RegularizedEvo ranked number one while Genetic CCN ranked the last. On the other hand, the accuracy ranking of all algorithms under the same experimental conditions in the proposed BenchENAS platform is: HierarchicalEvo \textgreater AECNN \textgreater CNN-GA \textgreater RegularizedEvo \textgreater CGP-CNN \textgreater NSGA-Net \textgreater Genetic-CNN \textgreater LargeEvo.
It can be seen that the performance of each ENAS algorithm under the same experimental conditions is not the same as quoted from the original papers. The comparability problem does exist in the comparison of the algorithms. Next, we will analyze why the situation that leads to different performance from the original paper occurs.

From the two rankings, we can observe that RegularizedEvo, NSGA-Net, and LargeEvo rankings are different from the ones in the original paper. The LargeEvo algorithm evolved from a linear structure in the original paper, starting with 1000 individuals per generation and evolving for 11 days to obtain a classification accuracy of 94.60$\%$. In this experiment, the LargeEvo algorithm stopped evolving after searching only 400 individuals. Therefore, the obtained neural network is a shallow one with poor performance. 
For NSGA-Net, we believe that it is the result of the combination of different data settings, different trainer settings, and different number of function evaluations. 
In the original paper of NSGA-Net, it incorporates a data preprocessing technique cutout~\cite{devries2017improved}, and a regularization technique, scheduled path~\cite{zoph2018learning}. To further improve the training process, an auxiliary head classifier is also appended to the DNN search.
To ensure the same setup for each experiment, these tricks are not applied in our experiments.
In addition, the number of function evaluations of NSGA-Net is 1,000 in the original paper, while the number is only 400 in this experiment.
These factors together cause NSGA-Net to perform less satisfactorily than the original paper showed.
For the RegularizedEvo algorithm, the number of function evaluations of the original paper is 20000 while the number is only 400 in this experiment. In addition, the RegularizedEvo algorithm uses the model augmentation trick in~\cite{zoph2018learning}, RMSProp optimizers and the scheduled path in training that are not used in this experiment. As a result, the performance is not as good as the original paper indicated. 

Therefore, we believe that BenchENAS is effective for fair comparisons of ENAS algorithms. Through the training results of this platform, users can analyze the advantages and disadvantages of each algorithm more objectively without being influenced by other conditions such as trainer settings, data settings, and the number of function evaluations.

\section{Conclusions and Future Work}
This paper has introduced a benchmarking platform for ENAS, named BenchENAS. Version 1.0 of BenchENAS includes nine EC-based NAS algorithms. BenchENAS is completely open-sourced, such that interested users are able to develop new algorithms on top of it.
In a nutshell, BenchENAS provides a platform for the fair comparisons of different ENAS algorithms. Furthermore, BenchENAS elaborates an evaluator to speed up the fitness evaluation and save much computational resource by the cache components and the parallel components. 
BenchENAS has easy extensibility and good usability so that users can easily extend their own algorithms and easily use the platform.
We have done comparative experiments on BenchENAS using eight state-of-the-art ENAS algorithms to demonstrate that the fair comparison issue is possible and to provide benchmark data for future studies by researchers.

However, there are still issues in BenchENAS to be attended to. Firstly, the number of implemented ENAS algorithms remains small. Secondly, the efficient evaluator designed in BenchENAS only alleviates the problem of expensive computational resources, but does not solve the problem completely. Finally, the search spaces of different ENAS algorithms are still different at present. As expected, there is still room for improvement for a truly fair comparison.

In the future, we will keep trying to solve the above problems. 
Firstly, although BenchENAS allows the users to submit their own codes to be included, we will keep following and adding more state-of-the-art ENAS algorithms into BenchENAS. The authors with promising ENAS algorithms will also be actively solicited. Secondly, some efficient methods for evaluating DNNs already exist to solve the problem of time-consuming DNN evaluation, and we will consider incorporating them into BenchENAS. Finally, we will exploit means to make all algorithms compare in the same search space.
Furthermore, we will continuously maintain and develop BenchENAS for years to come.

\ifCLASSOPTIONcaptionsoff
  \newpage
\fi

\bibliographystyle{IEEEtran}
\bibliography{reference}

\end{document}